\documentclass[a4paper]{article}

\usepackage{INTERSPEECH2021}
\usepackage{url}
\usepackage{hyperref}
\title{Semantic sentence similarity: size does not always matter}
\name{Danny Merkx, Stefan L. Frank, Mirjam Ernestus}

\address{
  Centre for Language Studies, Radboud University, Nijmegen, The Netherlands
  }
\email{danny.merkx@ru.nl,  s.frank@let.ru.nl, m.ernestus@let.ru.nl}

\begin{document}

\maketitle
\begin{abstract}

This study addresses the question whether visually grounded speech recognition (VGS) models learn to capture sentence semantics without access to any prior linguistic knowledge. We produce synthetic and natural spoken versions of a well known semantic textual similarity database and show that our VGS model produces embeddings that correlate well with human semantic similarity judgements. Our results show that a model trained on a small image-caption database outperforms two models trained on much larger databases, indicating that database size is not all that matters. We also investigate the importance of having multiple captions per image and find that this is indeed helpful even if the total number of images is lower, suggesting that paraphrasing is a valuable learning signal. While the general trend in the field is to create ever larger datasets to train models on, our findings indicate other characteristics of the database can just as important important. 

\end{abstract}
\noindent\textbf{Index Terms}: speech recognition, multimodal  embeddings, computational linguistics, deep learning, sentence semantics

\section{Introduction}

The idea that words that occur in similar contexts have similar meaning has been investigated for decades (e.g., \cite{Rubenstein65,Deerwester1990}). Advances in deep learning and computational power have made it possible to create models that learn useful and meaningful representations of larger spans of text such as sentences, paragraphs and even complete documents \cite{kiros2015,Hill2016,Conneau2017,Yang2018,Kiela2018,devlin2018}. A caveat of such models is the need to be trained on enormous amounts of text and the current trend is to use ever larger training corpora to create better models. Whereas BERT \cite{devlin2018} is trained on 2.5 billion tokens of text the more recent GPT-3 \cite{brown2020} is trained on nearly 500 billion tokens. It is obvious that humans are able to understand and use language after much less exposure; one would need to read 200 words per second, 24 hours a day for 80 years to digest as much information as GPT-3. People are able to hear and speak long before we are able to read, and many people never learn to read at all. Moreover, writing is a relatively recent invention, which only arose after spoken language.

Visually Grounded Speech (VGS) models aim to learn language without using written text data or prior information on the linguistic units in the speech signal. Instead, these models combine speech signals with visual information to guide learning; a VGS model learns to create representations for an image and its corresponding spoken caption that are similar to each other in the embedding space. Such models have been shown to learn to extract meaningful linguistic units from speech without explicitly being told what these units are, as shown in word recognition experiments \cite{merkx2019language,havard2019word} and semantic keyword spotting \cite{kamper2018visually}. Recent research has shown that VGS models with quantisation layers learn to extract phonetic and word-like units that are useful in zero-shot learning and speech synthesis \cite{HarwathVQ,Hsu2020}.  

As with text-based models, there is a trend in VGS models to use ever larger training corpora. CELL, one of the earliest VGS models, used a database of around 8,000 utterances \cite{Roy1998}. Harwath and colleagues introduced the first `modern' neural network based approach which was trained on the Flickr8k Audio Caption corpus, a corpus with 40,000 utterances \cite{Harwath2015}. This corpus was quickly followed up by Places Audio Captions (400,000 utterances) \cite{Harwath2018} and, most recently, by SpokenCOCO (600,000 utterances) \cite{Hsu2020}. 

However, previous work on visual grounding using written captions has shown that larger databases do not always result in better models. In \cite{merkx2019b}, we compared models trained on the written captions of Flickr8k \cite{Hodosh2015} and MSCOCO \cite{Chen2015}. We showed that, although the much larger MSCOCO (600k sentences) achieved better performance on the training task, the model trained on the smaller Flickr database performed better at transfer tasks; the resulting embeddings correlated better with human semantic relatedness ratings. As the MSCOCO model only performed better on visually descriptive sentences, these results suggest that there is a trade-off between getting better at processing image descriptions and creating generally useful sentence representations. 

There is another interesting difference between the VGS training corpora besides their size. While both Flickr8k Audio and SpokenCOCO have five captions per image, Places Audio has only one. Consequently, even though SpokenCOCO has more captions than Places, Places has 400,000 images while SpokenCOCO has only 120,000. 
The more fundamental difference is how models trained on Places and Flickr8k handle paraphrases. In a VGS, captions with similar images (i.e., likely paraphrases) should have similar representations. So, in a way, a VGS can be said to implicitly learn that paraphrases share one meaning. However, the paraphrasing in SpokenCOCO and Flickr8k is more explicit than in Places because there are always five captions per image, and these should ideally have the same representation in the embedding space.

Our first research question is: do VGS models learn to capture sentence semantics? So far, testing of the usability of the sentence representations created by VGS models has been limited, and recent research has focused more on whether useful sub-sentence units can be extracted (e.g. \cite{havard2019word,HarwathVQ,Hsu2020}). To answer this question we will investigate whether the representations learned by a VGS are predictive of semantic sentence similarity as judged by humans. In order to test this, we create spoken versions of the Semantic Textual Similarity (STS) database. STS consists of sentence pairs that were annotated by humans for semantic similarity. We look at the correlation between the human similarity ratings and the similarity of sentence representations created by our VGS model. 

We compare models trained on the three spoken image caption databases; Flickr8k Audio Captions, Places Audio Captions and SpokenCOCO. It is tempting to simply move on to the bigger corpus once one becomes available without investigating whether this actually constitutes an improvement. Using more data will likely lead to an increase in training task performance, but comparisons between corpora based on metrics other than training task performance are scarce. We investigate which model creates sentence representations that best capture sentence semantics, the only difference between these models being the database they were trained on. Our test material (STS) contains sentences from a wide range of domains, so a model needs to be able to generalise well to perform well on this task. 

We will also investigate the importance of paraphrasing in corpora having multiple captions for each image. Our second research question is: is it beneficial for VGS models to have multiple captions per image? We answer this question by training models on subsets of SpokenCOCO where we fix the total number of captions, but vary the number of captions per image and consequently the number of images in the training data. 

\section{Methods}

\subsection{Semantic similarity data}

For the semantic evaluation we use the Semantic Textual Similarity (STS) data. STS is a shared task hosted at the SemEval workshop. These datasets contain paired sentences from various sources labelled by humans with a similarity score between zero (`the two sentences are completely dissimilar') and five (`the two sentences are completely equivalent, as they mean the same thing') averaged over five annotators per sentence pair (see \cite{Agirre2015} for a full description of the annotator instructions). 

We use the STS 2012 to 2016 tasks, which are included in the SentEval toolbox for testing textual sentence representations \cite{Conneau2018}, allowing for a comparison between speech-based models and previous work using SentEval. Table \ref{STS} gives an overview of the STS tasks by year, and the sources from which the sentences were taken. We had the sentences produced by speech production software (synthetic speech) and by humans. All synthetic and natural utterances are made publicly available in .wav format as the SpokenSTS database. 

\subsubsection{Synthetic speech}

The synthetic speech was created with Google's Wavenet using three male and three female voices with a US accent. All utterances were produced using all six voices for a total of 75,264 utterance pairs. We applied as little preprocessing to the STS text as possible. To identify the necessary preprocessing steps, we sampled 10\% of the STS sentence pairs to convert to synthetic speech without any preprocessing. This sample was used to identify text characteristics that were troublesome to Wavenet and to apply the necessary preprocessing steps in order to correct these where possible. For example, Wavenet pronounces the quotation marks (saying `quote') if there is a space between the period and a quotation mark at the end of a sentence. Wavenet also pronounces certain non-capitalised abbreviations as if they were words rather than spelling them out (e.g., ``usa" is pronounced /usa/ instead of /u/, /s/, /a/). A full overview of all preprocessing applied, our code and our data can be found at \url{https://github.com/DannyMerkx/speech2image/tree/Interspeech21}. 

\subsubsection{Natural speech}

We selected a random sample of 5\% of the STS sentence pairs (638 pairs) evenly distributed across the different STS subsets. These sentences were recorded by four native speakers of English (two male, two female) with a North American accent. Recordings were made in a sound-attenuated booth using Audacity in sessions of one and a half hour including breaks. Speakers read the sentences out loud from a script. They were instructed to pronounce the sentences as they found most appropriate (e.g., saying `an apple' even though the original STS sentence might be misspelled as `a apple') and to pronounce large numbers according to their preference either in full or digit by digit. Speakers were paid 10 euros per hour in gift certificates. 

After recording was done, the audio was processed by an annotator. Utterances were automatically detected and labelled in Audacity, checked by the annotator for deviations from the script and where possible these deviations were corrected. For instance, when speakers made a mistake, they were allowed to continue from a natural break like a comma and so the annotator combined the correct parts from multiple attempts. If speakers misspoke and corrected themselves mid-utterance without re-recording (part of) the utterance, the mistake was removed. Furthermore, silences longer than 500ms were shortened. 

\begin{table}
    \caption{Description of the STS subtasks by year. Some subtasks appear in multiple years, but consist of different sentence pairs drawn from the same source. The image description datasets are drawn from the PASCAL VOC-2008 dataset \cite{pascal-voc-2008} and do not overlap with the training material of our models.}
        \resizebox{1\linewidth}{!}{
        \begin{tabular}{l l r l } 
        \toprule
         Task & Subtask & \#Pairs & Source \\ \hline
          & MSRpar & 750 & newswire\\
         & MSRvid & 750 & videos\\
         STS 2012 & SMTeuroparl & 459 & glosses\\
         & OnWN & 750 & WMT eval.\\
         & SMTnews & 399 & WMT eval.\\ \hline
         & FNWN & 189 & newswire\\
         STS 2013 & HDL & 750 & glosses\\
         & OnWN & 561 & glosses\\ \hline
         & Deft-forum & 450 & forum posts\\
         & Deft-news & 300 & news summary\\
         STS 2014 & HDL & 750 & newswire headlines\\
         & Images & 750 & image descriptions\\
         & OnWN & 750 & glosses\\
         & Tweet-news & 750 & tweet-news pairs\\ \hline
         & Answers forum & 375 & Q\&A forum answers\\
         & Answers students & 750 & student answers\\
         STS 2015 & Belief & 375 & committed belief\\
         & HDL & 750 & newswire headlines\\
         & Images & 750 & image descriptions\\ \hline
         & Answer-Answer & 254 & Q\&A forum answers\\
         & HDL & 249 & newswire headlines\\
         STS 2016 & Plagiarism & 230 & short-answer plagiarism\\
         & Postediting & 244 & MT postedits\\
         & Question-Question & 209 &  Q\&A forum questions\\ \hline
         Total & & 12,544 & \\
         \bottomrule
        \end{tabular}}
    \label{STS}
\end{table}

\subsection{Visually Grounded Speech Model}

The VGS architecture used in this study is our own implementation presented in \cite{merkx2019language} and we refer to that paper for more details. Here, we present a description of the model and the differences with \cite{merkx2019language}.

Our VGS model maps images and their corresponding captions to a common embedding space. It is trained to make matching images and captions lie close together, and mismatched images and captions lie far apart, in the embedding space. The model consists of two parts; an image encoder and a caption encoder. The image encoder is a single-layer linear projection on top of ResNet-152 \cite{He2015}, a pretrained image recognition network, with the classification layer removed. We train only the linear projection and do not further fine-tune ResNet.

The caption encoder consists of a 1-dimensional convolutional layer followed by a bi-directional recurrent layer and finally a self-attention layer. The only difference with \cite{merkx2019language} is the use of a four-layer LSTM instead of a three-layer GRU. Audio features consist of 13 Cepstral mean-variance normalised MFCCs and their first and second order derivatives calculated for 25ms frames with 10ms frameshift.  

\subsection{Training material}

We train separate models on each of the three training corpora. Flickr8k \cite{Hodosh2015} has 8,000 images and 40,000 written captions, five per image. We use the spoken versions of these captions collected using Amazon Mechanical Turk (AMT) by \cite{Harwath2015}. The data split is provided by \cite{Karpathy2017}, with 6,000 images for training and a development and test set both of 1,000 images. 

Places has 400,000 images drawn from the Places205 corpus \cite{Zhou2014} for which a single audio description per image was collected by \cite{Harwath2018} using AMT. Whereas Flickr8k Audio consisted of written captions which were then read out loud by workers, here, workers were tasked with describing the Places images as no written captions existed. We use the most recent official split\footnote{Available at: \url{https://groups.csail.mit.edu/sls/downloads/placesaudio/downloads.cgi}} with 400,000 images for training and a development and test set of 1,000 images.

MSCOCO has 123,287 images and 605,495 written captions \cite{Chen2015}, for which \cite{Hsu2020} collected spoken versions using AMT which they released as SpokenCOCO. Five thousand images are reserved as a development set and no official test set is provided. In order to keep results comparable between models we use 1,000 images from the development set for development and reserve 1,000 images as a test set.

\subsection{Experiments}

All models reported in this study are trained for 32 epochs. The models are trained using a cyclical learning rate which smoothly varies the learning rate between $2\times 10^{-4}$ and $2\times 10^{-6}$ over the course of four epochs. After a model is trained, we select the epoch with the lowest development set error for further testing. 

To answer our first research question, we use the trained caption encoders to encode the SpokenSTS sentences. We calculate the cosine similarity between each pair of encoded sentences and then calculate the Pearson correlation coefficient between the embedding similarity scores and the human similarity judgements. 

To answer our second research question, we train five more models on subsets of SpokenCOCO where we vary the number of images in the training set and the number of captions per image. As a lower bound on the amount of data we take the size of Flickr8k; 6,000 images and 30,000 captions, five per image. We then increase the amount of visual information (i.e., number of images) while keeping the total number of captions fixed at 30,000; 7,500 images with four captions per image, 10,000 images with three captions per image, 15,000 images with two captions per image and finally 30,000 images with one caption per image, similar to the Places database. If paraphrasing is helpful to the model, we expect model performance to decrease with a decreasing number of captions per image, even though the total number of captions remains the same. While we obviously cannot make sure that the models are trained on the same data, the data in the model with five captions per image is a subset of the data for the model with four captions per images and so on, so that the training data for each model overlaps as much as possible given the experimental setup. All code used in this study is available at \url{https://github.com/DannyMerkx/speech2image/tree/Interspeech21}

\section{Results}

In Table \ref{flickr_c2i_results} we compare the image-caption retrieval performance of the three models trained on different datasets (Flickr8k Audio, Places Audio and SpokenCOCO). This indicates how well the models perform on the training task. In order to retrieve images using a caption or captions using an image, the caption embeddings are ranked by their similarity to the image embeddings, and vice versa. It is clear that training task performance increases with database size.

\begin{figure*}[t]
    \centering
    \includegraphics[width=\linewidth]{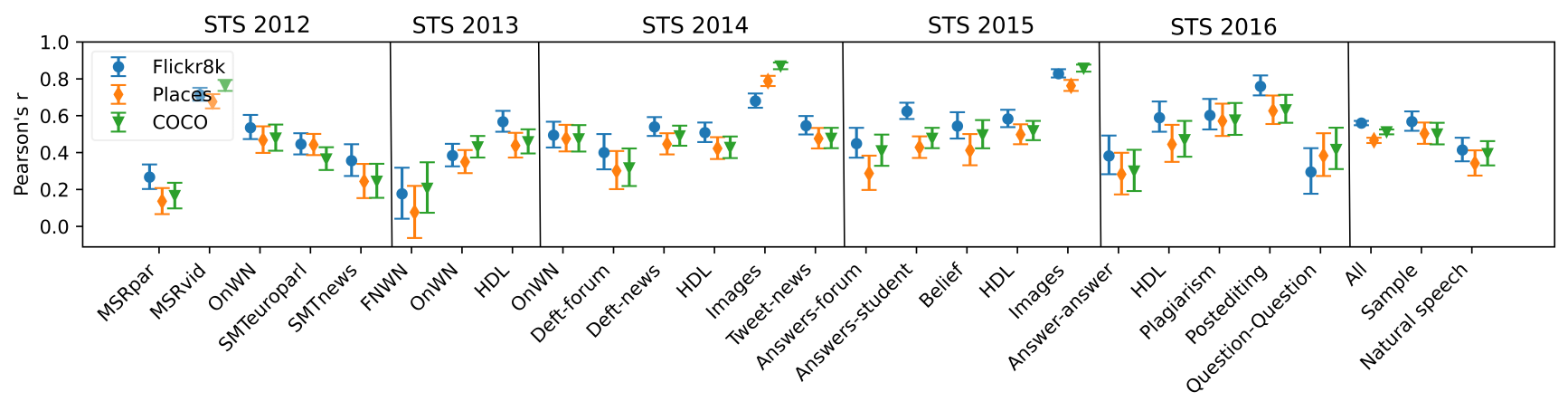}
    \caption{Semantic evaluation task results: Pearson correlation coefficients with their 95 percent confidence interval for the various subtasks using the synthetic SpokenSTS(see Table~\ref{STS}). The rightmost section shows the average over all STS subsets (All), the results on the natural speech recordings (Natural speech) and the results on the synthetic version of the natural speech sample (Sample).}
    \label{errorplots}
\end{figure*}

The results of the sentence semantics evaluation are shown in Figure \ref{errorplots}. We show Pearson correlation coefficients between the human similarity judgements and the embedding similarities generated by the trained models. As each sentence is pronounced by six voices we calculate the embedding similarity for each pair of voices and average over the resulting 36 pairs. In general, we see that both the Flickr8k and the SpokenCOCO model tend to outperform the Places model, and that the Flickr8k model tends to outperform the SpokenCOCO model. This is confirmed by the significant differences in Pearson's r calculated on the complete SpokenSTS database (indicated as All).

Lastly, it is clear that all models perform worse on natural speech. In Figure \ref{errorplots}, Sample indicates the subset of synthetic speech representing the same sample of STS sentences that was used for the natural speech. Model performance on this subset is similar to the performance on the entirety of SpokenSTS indicating that the sample is representative of the entire corpus.

\begin{figure}[t]
    \centering
    \includegraphics[width=.75\linewidth]{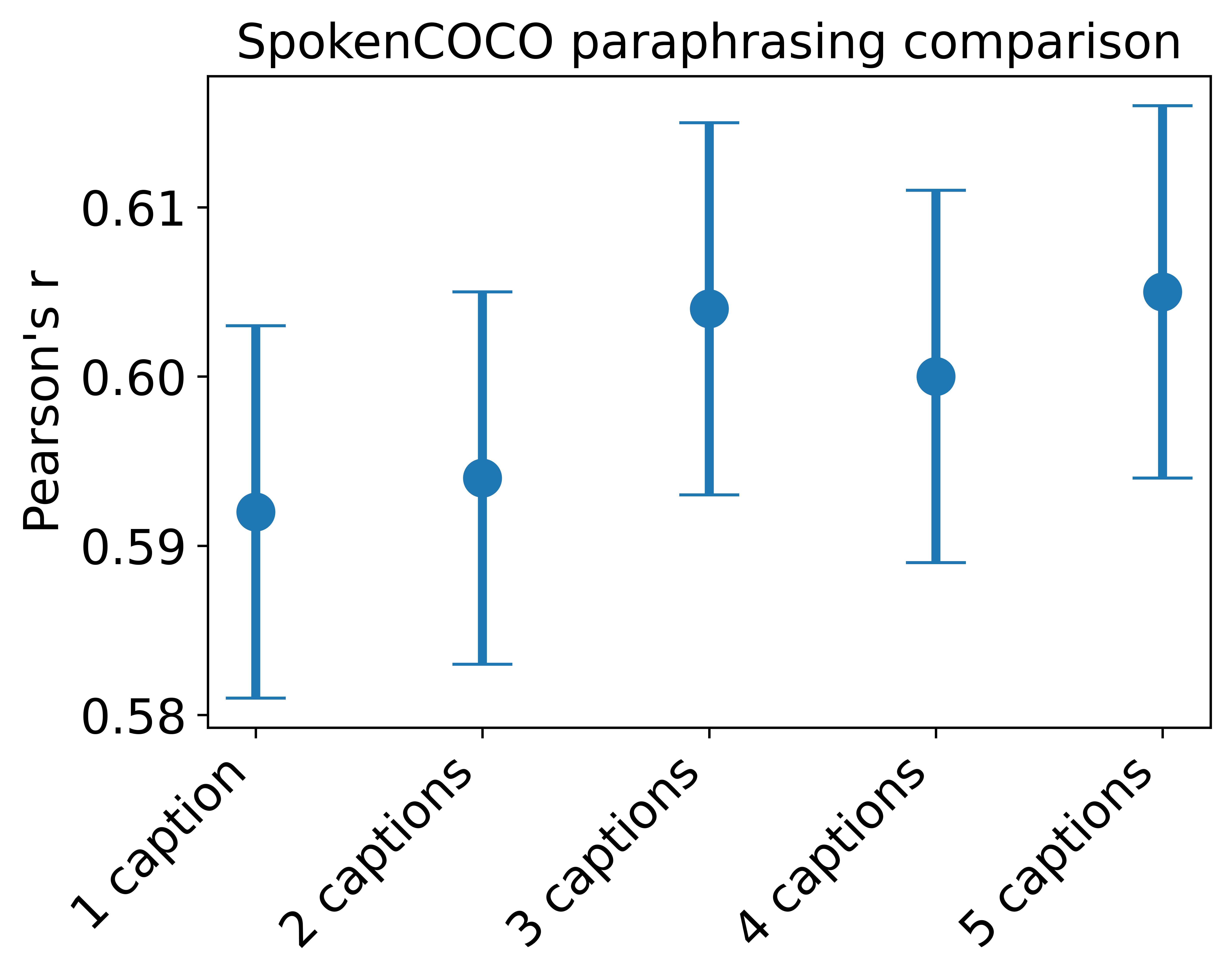}
    \caption{Comparison of the five models trained on subsets of SpokenCOCO with differing numbers of captions per image. We show Pearson correlation coefficients over the entire synthetic SpokenSTS with 95 percent confidence intervals.}
    \label{paraphrasing}
\end{figure}

To further investigate this trend we performed five separate regression analyses with the human similarity judgements as dependent variable and each of the five models' similarity ratings as regressors. Embedding similarities were averaged over the 36 voice pairs. Table \ref{aic} shows a comparison of the Akaike Information Criteria (AIC) of these regression models. These results show the same trend as Figure \ref{paraphrasing} and clearly indicate that the similarity ratings generated by models with more captions per image provide a better fit to the human similarity ratings. 

\begin{table}[!t]
    \caption{Image-Caption retrieval results of each database's respective test set. R@N is the percentage of items for which the correct image or caption was retrieved in the top N (higher is better). Med r is the median rank of the correct image or caption (lower is better).}
    \centering
        \begin{tabular}{l | r r r r }
            \toprule
            \multicolumn{1}{l}{Model} & \multicolumn{4}{c}{Caption to Image} \\ \hline

             & R@1 & R@5 & R@10 & med r \\ \hline
            Flickr8k Audio & 12.7 & 35.1 & 48.4 & 12 \\ 
            Places Audio & 30.6 & 62.6 & 73.8 & 3 \\ 
            SpokenCOCO & 30.6 & 64.1 & 79.8& 3 \\ 
             & \multicolumn{4}{c}{Image to Caption} \\ \hline
            Flickr8k Audio & 20.3 & 44.8 & 58.8 & 7 \\ 
            Places Audio & 29.5 & 62.0 & 74.3 & 3 \\ 
            SpokenCOCO & 39.2 & 75.3 & 86.4 & 2 \\ 
            \bottomrule
        \end{tabular}
    \label{flickr_c2i_results}
\end{table}

\begin{table}
    \caption{AIC comparison of regression models (lower is better). $\Delta$AIC indicates the difference in AIC compared to the best model, LL indicates the model's log likelihood}
    \centering
        \begin{tabular}{l | r r r }
            \toprule
            No. captions & AIC & $\Delta$AIC & LL \\ \hline
            5 &  127974.5 & 0.00  & $-$63984.23 \\
            3 & 127985.3 & 10.81  & $-$63989.64\\
            4 &  128116.3 & 141.80  & $-$64055.13\\
            2 & 128218.3 & 243.80  & $-$64106.13\\
            1 &  128269.7 & 295.26  & $-$64131.86\\
            \bottomrule
        \end{tabular}
    \label{aic}
\end{table}

\section{Discussion and conclusion}

We collected synthetic and natural speech for a large corpus of human sentence similarity judgements in order to investigate whether VGS models learn to capture sentence semantics. Furthermore, we investigated the merits of database size and the availability of paraphrases in the training data. 

The results show that similarity scores generated by our VGS models correlate quite well human similarity judgements overall. This shows that a model tasked with mapping images to captions and vice versa learns to capture sentence semantics. However there are also some subsets of STS (MSRpar, FNWN) on which the model performs quite poorly, and unsurprisingly all models clearly perform best on subtasks consisting of visual descriptions. Furthermore, we found that even though the models trained on Places and SpokenCOCO outperform the Flickr8k model in terms of training task performance, the Flickr8k model performs better on the SpokenSTS task. This confirms previous results on text-based grounding models \cite{merkx2019b} which compared models trained on the written versions of Flickr8k and MSCOCO. As in \cite{merkx2019b} we see that SpokenCOCO outperforms Flickr8k mainly on the subtasks containing visual descriptions. The models that were trained on a smaller subsets of SpokenCOCO for the paraphrasing experiment performed better than the model trained on the entire database. This indicates that training on more data might cause the model to overspecialise; it performs better on sentences which are similar to the training data, but becomes worse at generalising to sentences from other domains. 

Next, we investigated whether the presence of paraphrases in the data (i.e., multiple captions per image) is beneficial to the model. By training models on subsets of SpokenCOCO where we fixed the total number of captions but varied the number of captions per image, we found that having more captions per image increases model performance, even though these models consequently are trained on less visual information. This also explains why the Places model performs worst out of the three, even though the amount of data is in the same ballpark as SpokenCOCO (it has fewer captions but more images). An interesting question for future research is whether this trend continues beyond five captions per image. Collecting more captions for existing databases, rather than collecting more image captions pairs, could be an important consideration for future data collection efforts. 

In conclusion, we show VGS models are capable of capturing sentence semantics. Importantly, our results show that database size is not all that matters when it comes to training VGS models. Even though it is enticing to collect ever larger databases to increase training task performance, this does not always translate to better transfer learning results. Our Flickr8k model outperforms our SpokenCOCO model even though it has 20 times less data. Furthermore, other characteristics of a database might be even more important than its size; in the case of VGS this is the presence of multiple captions per image. 

\section{Acknowledgements}

The research presented here was funded by the Netherlands Organisation for Scientific Research (NWO) Gravitation Grant 024.001.006 to the Language in Interaction Consortium.

\bibliographystyle{IEEEtran}

\bibliography{mybib}

\begin{thebibliography}{10}
\providecommand{\url}[1]{#1}
\csname url@samestyle\endcsname
\providecommand{\newblock}{\relax}
\providecommand{\bibinfo}[2]{#2}
\providecommand{\BIBentrySTDinterwordspacing}{\spaceskip=0pt\relax}
\providecommand{\BIBentryALTinterwordstretchfactor}{4}
\providecommand{\BIBentryALTinterwordspacing}{\spaceskip=\fontdimen2\font plus
\BIBentryALTinterwordstretchfactor\fontdimen3\font minus
  \fontdimen4\font\relax}
\providecommand{\BIBforeignlanguage}[2]{{%
\expandafter\ifx\csname l@#1\endcsname\relax
\typeout{** WARNING: IEEEtran.bst: No hyphenation pattern has been}%
\typeout{** loaded for the language `#1'. Using the pattern for}%
\typeout{** the default language instead.}%
\else
\language=\csname l@#1\endcsname
\fi
#2}}
\providecommand{\BIBdecl}{\relax}
\BIBdecl

\bibitem{Rubenstein65}
H.~Rubenstein and J.~B. Goodenough, ``Contextual correlates of synonymy,''
  \emph{Communications of the Association for Computing Machinery}, vol.~8,
  no.~10, pp. 627--633, 1965.

\bibitem{Deerwester1990}
S.~Deerwester, S.~T. Dumais, G.~W. Furnas, T.~K. Landauer, and R.~Harshman,
  ``Indexing by latent semantic analysis,'' \emph{Journal of the American
  Society for Information Science}, vol.~41, no.~6, pp. 391--407, 1990.

\bibitem{kiros2015}
R.~Kiros, Y.~Zhu, R.~R. Salakhutdinov, R.~Zemel, R.~Urtasun, A.~Torralba, and
  S.~Fidler, ``Skip-thought vectors,'' in \emph{Advances in Neural Information
  Processing Systems 28 (NIPS)}, 2015, pp. 3294--3302.

\bibitem{Hill2016}
F.~Hill, K.~Cho, and A.~Korhonen, ``Learning distributed representations of
  sentences from unlabelled data,'' in \emph{Proceedings of the 2016 Conference
  of the North {A}merican Chapter of the Association for Computational
  Linguistics: Human Language Technologies (NAACL-HLT)}, 2016, pp. 1367--1377.

\bibitem{Conneau2017}
A.~Conneau, D.~Kiela, H.~Schwenk, L.~Barrault, and A.~Bordes, ``{Supervised
  Learning of Universal Sentence Representations from Natural Language
  Inference Data},'' in \emph{Proceedings of the 2017 Conference on Empirical
  Methods in Natural Language Processing (EMNLP)}, 2017, pp. 670--680.

\bibitem{Yang2018}
Y.~Yang, S.~Yuan, D.~Cer, S.-y. Kong, N.~Constant, P.~Pilar, H.~Ge, Y.-H. Sung,
  B.~Strope, and R.~Kurzweil, ``Learning semantic textual similarity from
  conversations,'' in \emph{Proceedings of The Third Workshop on Representation
  Learning for {NLP}}, 2018, pp. 164--174.

\bibitem{Kiela2018}
D.~Kiela, A.~Conneau, A.~Jabri, and M.~Nickel, ``Learning visually grounded
  sentence representations,'' in \emph{Proceedings of the 2018 Conference of
  the North {A}merican Chapter of the Association for Computational
  Linguistics: Human Language Technologies}, 2018, pp. 408--418.

\bibitem{devlin2018}
J.~Devlin, M.-W. Chang, K.~Lee, and K.~Toutanova, ``Bert: Pre-training of deep
  bidirectional transformers for language understanding,'' in \emph{Proceedings
  of the 2019 Conference of the North {A}merican Chapter of the Association for
  Computational Linguistics: Human Language Technologies (NAACL-HLT)}, 2019, p.
  4171–4186.

\bibitem{brown2020}
T.~B. Brown, B.~Mann, N.~Ryder, M.~Subbiah, J.~Kaplan, P.~Dhariwal,
  A.~Neelakantan, P.~Shyam, G.~Sastry, A.~Askell, S.~Agarwal, A.~Herbert-Voss,
  G.~Krueger, T.~Henighan, R.~Child, A.~Ramesh, D.~M. Ziegler, J.~Wu,
  C.~Winter, C.~Hesse, M.~Chen, E.~Sigler, M.~Litwin, S.~Gray, B.~Chess,
  J.~Clark, C.~Berner, S.~McCandlish, A.~Radford, I.~Sutskever, and D.~Amodei,
  ``Language models are few-shot learners,'' in \emph{34th Conference on Neural
  Information Processing Systems (NeurIPS 2020)}, 2020.

\bibitem{merkx2019language}
D.~Merkx, S.~L. Frank, and M.~Ernestus, ``Language learning using speech to
  image retrieval,'' in \emph{Proceedings of Interspeech 2019. Crossroads of
  Speech and Language}, 2019.

\bibitem{havard2019word}
W.~N. Havard, J.-P. Chevrot, and L.~Besacier, ``Word recognition, competition,
  and activation in a model of visually grounded speech,'' in \emph{Proceedings
  of the 23rd Conference on Computational Natural Language Learning (CoNLL)},
  2019, pp. 339--348.

\bibitem{kamper2018visually}
H.~Kamper and M.~Roth, ``Visually grounded cross-lingual keyword spotting in
  speech,'' in \emph{The 6th International Workshop on Spoken Language
  Technologies for Under-Resourced Languages}, 2018.

\bibitem{HarwathVQ}
D.~Harwath, W.-N. Hsu, and J.~Glass, ``Learning hierarchical discrete
  linguistic units from visually-grounded speech,'' in \emph{{ICLR} 2020 The
  Ninth International Conference on Learning Representations}, 2020.

\bibitem{Hsu2020}
W.-N. Hsu, D.~Harwath, C.~Song, and J.~Glass, ``Text-free image-to-speech
  synthesis using learned segmental units,'' in \emph{Thirty-fourth Conference
  on Neural Information Processing Systems (NeurIPS)}, 2020.

\bibitem{Roy1998}
D.~Roy and A.~Pentland, ``{Learning words from natural audio-visual input},''
  vol.~4, no.~1, 1998, pp. 1279--1282.

\bibitem{Harwath2015}
D.~Harwath and J.~Glass, ``Deep multimodal semantic embeddings for speech and
  images,'' in \emph{2015 IEEE Workshop on Automatic Speech Recognition and
  Understanding (ASRU)}, 2015, pp. 237--244.

\bibitem{Harwath2018}
D.~Harwath, A.~Recasens, D.~Sur{\'{i}}s, G.~Chuang, A.~Torralba, and J.~Glass,
  ``{Jointly discovering visual objects and spoken words from raw sensory
  input},'' \emph{International Journal of Computer Vision}, vol. 128, pp.
  620--641, 2020.

\bibitem{merkx2019b}
D.~Merkx and S.~L. Frank, ``Learning semantic sentence representations from
  visually grounded language without lexical knowledge,'' \emph{Natural
  Language Engineering}, vol.~25, no.~4, p. 451–466, 2019.

\bibitem{Hodosh2015}
M.~Hodosh, P.~Young, and J.~Hockenmaier, ``Framing image description as a
  ranking task: Data, models and evaluation metrics,'' \emph{Journal of
  Artificial Intelligence Research}, vol.~47, no.~1, pp. 853--899, 2013.

\bibitem{Chen2015}
X.~Chen, H.~Fang, T.-Y. Lin, R.~Vedantam, S.~Gupta, P.~Dollar, and C.~L.
  Zitnick, ``{Microsoft COCO Captions: Data Collection and Evaluation
  Server},'' \emph{arXiv preprint arXiv: 1504.00325}, pp. 1--7, 2015.

\bibitem{Agirre2015}
E.~Agirre, C.~Banea, C.~Cardie, D.~Cer, M.~Diab, A.~Gonzalez-Agirre, W.~Guo,
  I.~Lopez-Gazpio, M.~Maritxalar, R.~Mihalcea, G.~Rigau, L.~Uria, and J.~Wiebe,
  ``{SemEval-2015 Task 2: Semantic Textual Similarity, English, Spanish and
  Pilot on Interpretability},'' in \emph{SemEval}, 2015, pp. 252--263.

\bibitem{Conneau2018}
A.~Conneau and D.~Kiela, ``{SentEval: An Evaluation Toolkit for Universal
  Sentence Representations},'' in \emph{Proceedings of the Eleventh
  International Conference on Language Resources and Evaluation ({LREC} 2018)},
  2018.

\bibitem{pascal-voc-2008}
M.~Everingham, L.~Van~Gool, C.~K.~I. Williams, J.~Winn, and A.~Zisserman, ``The
  {PASCAL} {V}isual {O}bject {C}lasses {C}hallenge 2008 {(VOC2008)}
  {R}esults,''
  http://www.pascal-network.org/challenges/VOC/voc2008/workshop/index.html,
  2008.

\bibitem{He2015}
K.~He, X.~Zhang, S.~Ren, and J.~Sun, ``Deep residual learning for image
  recognition,'' in \emph{The IEEE Conference on Computer Vision and Pattern
  Recognition (CVPR)}, 2016.

\bibitem{Karpathy2017}
A.~Karpathy and L.~Fei-Fei, ``Deep visual-semantic alignments for generating
  image descriptions,'' in \emph{The IEEE Conference on Computer Vision and
  Pattern Recognition (CVPR)}, 2015.

\bibitem{Zhou2014}
B.~Zhou, A.~Lapedriza, J.~Xiao, A.~Torralba, and A.~Oliva, ``Learning deep
  features for scene recognition using {Places} database,'' in \emph{Advances
  in Neural Information Processing Systems 27 (NIPS)}, 2014.

\end{thebibliography}


\end{document}